\documentclass[letterpaper]{article} 
\usepackage[preprint]{aaai24}  
\usepackage{times}  
\usepackage{helvet}  
\usepackage{courier}  
\usepackage[hyphens]{url}  
\usepackage{graphicx} 
\usepackage{natbib}  
\usepackage{caption} 
\usepackage{algorithm}
\usepackage{algorithmic}

\usepackage{newfloat}
\usepackage{listings}
\DeclareCaptionStyle{ruled}{labelfont=normalfont,labelsep=colon,strut=off} 
\floatstyle{ruled}
\newfloat{listing}{tb}{lst}{}
\floatname{listing}{Listing}
\title{{AUC} Optimization from Multiple Unlabeled Datasets}
\author {
    Zheng Xie$^*$,
    Yu Liu$^*$,
    Ming Li
}
\affiliations {
    National Key Laboratory for Novel Software Technology,\\ Nanjing University, Nanjing 210023, China\\
    \{xiez, liuy, lim\}@lamda.nju.edu.cn
}

\usepackage{amsmath}
\usepackage{amsthm}

\newtheorem{theorem}{Theorem}

\DeclareMathOperator{\auc}{AUC}

\DeclareMathOperator{\tpr}{TPR}
\DeclareMathOperator{\fpr}{FPR}

\DeclareMathOperator*{\Ebb}{\mathbb{E}}
\newcommand{\Ibb}{\mathbb{I}}

\newcommand{\elloi}{\ell_{01}}

\newcommand{\simiid}{\stackrel{\mathrm{i.i.d.}}{\sim}}

\newcommand{\Rcal}{\mathcal{R}}
\newcommand{\Xcal}{\mathcal{X}}

\newcommand{\Xcalp}{\mathcal{X}_{P}}
\newcommand{\Xcaln}{\mathcal{X}_{N}}

\newcommand{\Xcall}{\mathcal{X}_{L}}

\newcommand{\risk}[1]{R_{{#1}}}
\newcommand{\emprisk}[1]{\hat{R}_{{#1}}}

\newcommand{\Rpn}{\risk{PN}}
\newcommand{\empRpn}{\emprisk{PN}}

\newcommand{\Rij}{\risk{ij}}
\newcommand{\empRij}{\emprisk{ij}}
\newcommand{\Rum}{R_{U^m}}
\newcommand{\empRum}{\hat R_{U^m}}

\newcommand{\xv}{\boldsymbol{x}}

\newcommand{\xp}{\boldsymbol{x}}

\newcommand{\xn}{\boldsymbol{x}'}

\newcommand{\xpi}{\boldsymbol{x}_i}
\newcommand{\xnj}{\boldsymbol{x}'_j}

\newcommand{\EP}{\Ebb_{\xp \sim p_P (\xv)}}

\newcommand{\EN}{\Ebb_{\xn \sim p_N (\xv)}}

\newcommand{\Ei}{\Ebb_{\xp \sim p_i (\xv)}}

\newcommand{\Ej}{\Ebb_{\xn \sim p_j (\xv)}}

\newcommand{\citeay}[1]{\citeauthor{#1}~\shortcite{#1}}

\newcommand{\ie}{i.e., }

\newcommand{\um}{U\(^m\){}}
\newcommand{\umauc}{U\(^m\)-AUC{}}
\graphicspath{{figure}}
\usepackage{cleveref}
\usepackage{nameref}
\usepackage{amsfonts}
\usepackage{booktabs}
\usepackage{multirow}

\begin{document}

\maketitle

\def\thefootnote{*}\footnotetext{Equal contribution.}\def\thefootnote{\arabic{footnote}}

\begin{abstract}
  Weakly supervised learning aims to make machine learning more powerful when the perfect supervision is unavailable, and has attracted much attention from researchers. Among the various scenarios of weak supervision, one of the most challenging cases is learning from multiple unlabeled (U) datasets with only a little knowledge of the class priors, or \um{} learning for short. In this paper, we study the problem of building an AUC (area under ROC curve) optimal model from multiple unlabeled datasets, which maximizes the pairwise ranking ability of the classifier. We propose \umauc{}, an AUC optimization approach that converts the \um{} data into a multi-label AUC optimization problem, and can be trained efficiently. We show that the proposed \umauc{} is effective theoretically and empirically.
\end{abstract}

\section{Introduction}

Since obtaining perfect supervision is usually challenging in the real-world machine learning problems, the machine learning approaches often have to deal with inaccurate, incomplete, or inexact supervisions, collectively referred to as weak supervision~\cite{wsl}. To achieve this, many researchers have devoted into the area of \emph{weakly supervised learning}, such as semi-supervised learning~\cite{zhubook}, positive-unlabeled learning~\cite{pulsurvey}, noisy label learning~\cite{Han2021nllsurvey}, etc.

Among multiple scenarios of weakly supervised learning, one of the most challenging scenarios is to learn classifiers from \(m\) unlabeled (U) datasets with different class priors, i.e., the proportions of positive instances in the sets. Such a learning task is usually referred to as \um{} learning. This scenario usually occur when the instances can be categorized into different groups, and the probability of an instance to be positive varies across the groups, e.g., for predicting voting rates or morbidity rates. Prior studies include \citeay{scott20}, which ensembles the classifiers trained on all pairs of the unlabeled sets; \citeay{tsai20}, which introduces consistency regularization for the problem. 
Recently, \citeay{Lu2021umssc} proposed a consistent approach for classification from multiple unlabeled sets, which is the first classifier-consistent approach for learning from \(m\) unlabeled sets \((m>2)\) that optimizes a classification loss.

In this paper, we further consider the problem of learning an AUC (area under ROC curve) optimization model from the \um{} data, which maximizes the pairwise ranking ability of the classifier~\cite{auc}. The importance of this problem lie in two folds: First, we note that for certain scenarios, the ranking performance of the model is more concerned. E.g., ranking items with coarse-grind rank labels. Second, given multiple U sets with different class priors, the imbalance issue is very likely to affect the learning process. Thus, taking an imbalance-aware performance measure, i.e., AUC, is naturally appropriate for the problem.

To achieve this goal, we introduce \umauc{}, a novel AUC optimization approach from \um{} data. \umauc{} solves the problem as a multi-label AUC optimization problem, as each label of the multi-label learning problem corresponds to a pseudo binary AUC optimization sub-problem. To overcome the quadratic time complexity of the pairwise loss computation, we convert the problem into a stochastic saddle point problem and solve it through point-wise AUC optimization algorithm. Our theoretical analysis shows that \umauc{} is consistent with the optimal AUC optimization model, and provides the generalization bound. Experiments show that our approach outperforms the state-of-the-art methods and has superior robustness.

Our main contributions are highlighted as follows:
\begin{itemize}
    \item To the best of our knowledge, we present the first algorithm for optimizing AUC in \um{} scenarios. Additionally, our algorithm is the first to address the \um{} problem without the need for an exact class priors. Significantly, our algorithm possesses a simple form and exhibits efficient performance.

    \item Furthermore, we conduct a comprehensive theoretical analysis of the proposed methodology, demonstrating its validity and assessing its excess risk.
    
    \item We perform experiments on various settings using multiple benchmark datasets, and the results demonstrate that our proposed method consistently outperforms the state-of-the-art methods and performs robustly under different imbalance settings.
    
\end{itemize}

The reminder of our paper is organized as follows. We first introduce preliminary in \cref{sec:pre}. Then, we introduce the \umauc{} approach and conducts the theoretical analysis in \cref{sec:method}, while \cref{sec:exp} shows the experimental results. Finally,  \cref{sec:rela} introduces the related works, and \cref{sec:con} concludes the paper.

\section{Preliminary}\label{sec:pre}

In the fully supervised AUC optimization, we are given a dataset sampled from a specific distribution
\begin{equation}
    \Xcall := {\{(\xv_i, y_i) \}}_{i=1}^{n}
	\simiid p(\xv, y)\,. 
\end{equation}
For convenience, we refer to positive and negative data as samples from two particular distributions:
\begin{equation}
    \begin{aligned}
    \Xcalp :=& {\{\xpi \}}_{i=1}^{n_P}
    	\simiid
    	p_P(\xv):=p(\xv\mid y=+1)\,, \text{\;and} \notag \\
    \Xcaln :=& {\{\xnj \}}_{j=1}^{n_N}
    	\simiid
    	p_N(\xv):=p(\xv\mid y=-1)\,,
    \end{aligned}
\end{equation}
there we have \(\Xcall = \Xcalp \cup \Xcaln\).

Let \(f:\Xcal \rightarrow \Rcal\) be a scoring function. It is expected that positive instances will have a higher score than negative ones. For a threshold value \(t\), we define the true positive rate \(\tpr_f(t)=\Pr(f(x)\ge t | y = 1)\) and the false positive rate \(\fpr_f(t)=\Pr(f(x)\ge t | y = 0)\).
The AUC is defined as the area under the ROC curve:
\begin{equation}
\auc = \int_0^1 \tpr_f(\fpr^{-1}_f(t)) dt.
\end{equation}

Previous study~\cite{auc} introduced that randomly drawing a positive instance and a negative instance, the AUC is equivalent to the probability of the positive instance is scored higher than the negative instance, so that the AUC of the model \(f\) can be formulated as:
\begin{equation}
\auc = 1-\EP[\EN[\elloi(f(\xp) - f(\xn))]]\,.\label{eq:auc}
\end{equation}
Here \(\elloi(z)=\Ibb[z<0]\).
Without creating ambiguity, we will denote \(f(\xp, \xn)\) as \(f(\xp) - f(\xn)\) for clarity.

The maximization of the AUC is equivalent to the minimization of the following AUC risk. Since the true AUC risk measures the error rate of ranking positive instances over negative instances, we refer to the true AUC risk as the PN-AUC risk to avoid confusion:
\begin{equation}
\Rpn(f) = \EP\left[\EN[\elloi(f(\xp, \xn))]\right]\,.\label{eq:rpn}
\end{equation}

With a finite sample, we typically solve the following empirical risk minimization (ERM) problem:
\begin{equation}\label{eq:estimator:pn}
\min_f \quad \empRpn(f)=\frac{1}{|\Xcalp||\Xcaln|} \sum_{\xp \in \Xcalp}\sum_{\xn \in \Xcaln}\ell(f(\xp, \xn))\,.
\end{equation}

\section{U{$^m$}-AUC: The Method}\label{sec:method}

\begin{figure*}[t]
    \centering
    \includegraphics[width=\textwidth]{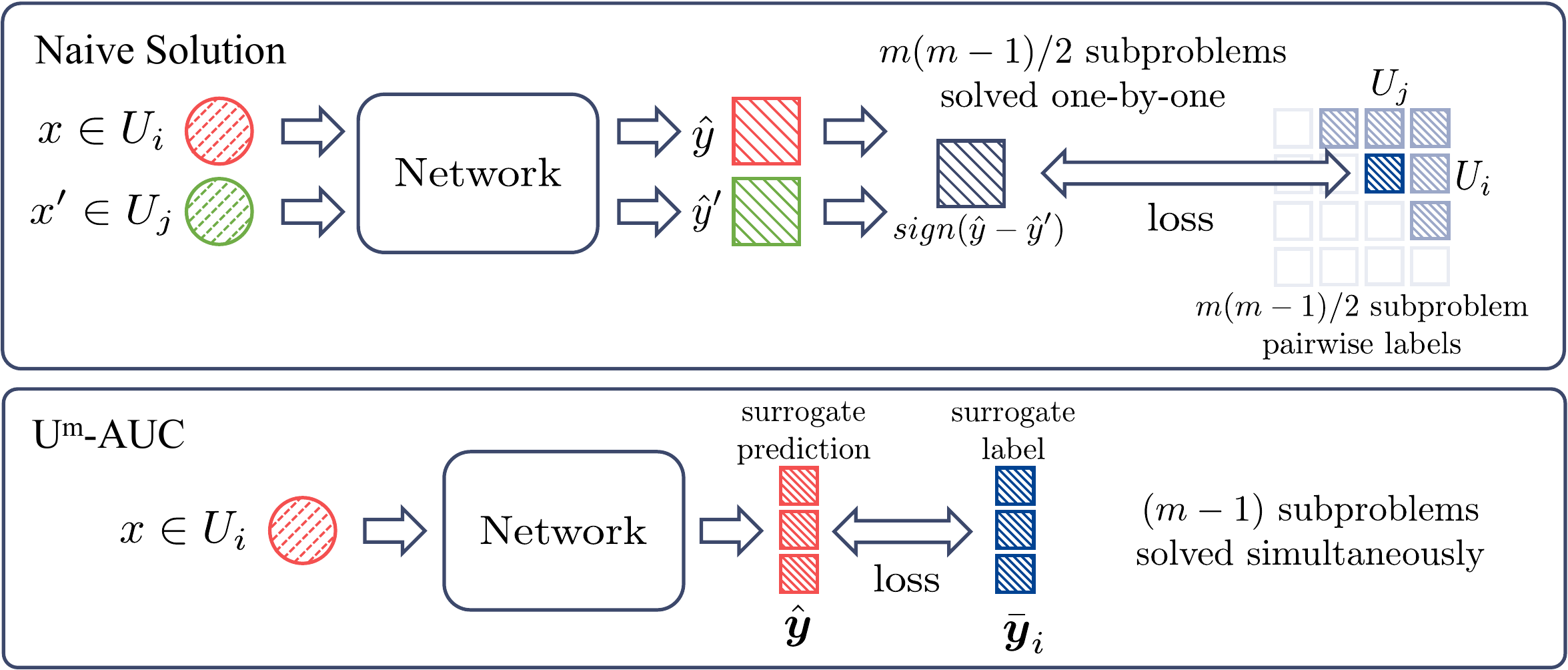}
    \caption{Framework Demonstration. The U{$^m$}-AUC employs an efficient stochastic optimization algorithm, eliminating the need for pairwise loss and reducing the time complexity to $O(n)$. Additionally, it simplifies the naive solution by transforming it into a multi-label learning problem, reducing the number of sub-problems and resulting in a more concise problem formulation.}
    \label{fig:framework}
\end{figure*}

In this paper, we study AUC optimization under U$^m$ setting, which involves optimizing AUC across multiple unlabeled datasets. Suppose we are given \(m(m \ge 2)\) unlabeled datasets \(\text{U}_1, \ldots, \text{U}_m\) with different class prior probabilities, as defined by the following equation:
\begin{equation}\label{umdef}
  \text{U}_i = \{x_{ik}\}_{k=1}^{n_i} \simiid p_i(x)= \pi_i p_P(x) + (1-\pi_i)p_N(x)\,,
\end{equation}
where \(p_P(x)\) and \(p_N(x)\) are the positive and negative class-conditional probability distributions, respectively, and \(\pi_i\) denotes the class prior of the \(i\)-th unlabeled set. The size of U$_i$ is $n_i$.
Although we only have access to unlabeled data, our objective is to build a classifier that minimizes the PN-AUC risk~\cref{eq:rpn}.

To achieve this goal, we propose \umauc{}, a novel AUC optimization approach that learns from \um{} data. Dislike the previous studies on \um{} classification who require the knowledge of the class prior~\cite{Lu2021umssc}, \umauc{} replace this requirement by only knowing knowledge of the relative order of the unlabeled sets based on their class priors, which is more realistic.

We next introduce the \umauc{} approach.
For convenience, and without loss of generality, we assume that the class priors of the unlabeled sets are in descending order, \ie \(\pi_i \geq \pi_j\) for \(i < j\). Additionally, we assume that at least two unlabeled sets have different priors, \ie \(\pi_1 > \pi_m\); otherwise, the problem is unsolvable.

\subsection{Consistent AUC Learning from U{$^m$} Data}

To provide a solution that is consistent with the true AUC, we first introduce the two sets case, i.e., one can achieve consistent AUC learning through two unlabeled sets with different class priors. Such a result is discussed in previous studies~\cite{Charoenphakdee2019}.
Suppose the two unlabeled sets are \(\text{U}_i\) and \(\text{U}_j\) with \(\pi_i > \pi_j\). We can minimize the following U\(^2\) AUC risk:
\begin{equation}
R_{ij}(f) = \Ei\left[\Ej[\elloi(f(\xp, \xn))]\right]\,
\end{equation}\label{eq:rtwosets}
by solve the following U\(^2\) AUC ERM problem:
\begin{equation}\label{eq:twosets}
  \min_f \quad \empRij(f)=\frac{1}{n_i n_j} \sum_{\xp \in \text{U}_i}\sum_{\xn \in \text{U}_j}\ell(f(\xp, \xn))\,.
\end{equation}

The following theorem shows that the U\(^2\) AUC risk minimization problem is consistent with the original AUC optimization problem we need to solve, \ie we can solve the original AUC optimization problem by minimizing the U\(^2\) AUC risk.

\begin{theorem}[U\(^2\) AUC consistency]
  Suppose $f^*$ is a minimizer of the AUC risk $\Rij$ over two distributions $p_i$ and $p_j$ where $\pi_i > \pi_j$,  \ie $f^* = \arg \min \Rij$.
        Then, it follows that $f^*$ is also a minimizer of the true AUC risk $\Rpn$ , and thus, $\Rij$ is consistent with $\Rpn$.
\end{theorem}

Therefore, by minimizing the U$^2$ AUC risk under the condition that only impure data sets are available, we can obtain the desired model.

With the \um{} data, we can construct the following minimization problem by composing \(m(m-1)/2\) AUC sub-problems with weights \(z_{ij} > 0\):
\begin{equation}\label{eq:rmsets}
\min_f \quad \Rum(f) = \sum_{i,j|1\le i < j \le m} z_{ij} \Rij(f)
\end{equation}
which corresponds to the U\(^m\) AUC ERM problem

\begin{equation}\label{eq:msets}
  \min_f \quad \empRum(f) = \sum_{i,j|1\le i < j \le m}   \sum_{\xp \in \text{U}_i}\sum_{\xn \in \text{U}_j}\frac{z_{ij}\ell(f(\xp, \xn))}{n_i n_j}\,.
\end{equation}

As well, we can theoretically demonstrate the consistency between U\(^m\) AUC risk minimization problem and the original AUC optimization problem.
\begin{theorem}[U$^m$ AUC consistency]\label{thm:con_msets}
  Suppose $f^*$ is a minimizer of the AUC risk $\Rum$ over $m$ distributions $p_1, \cdots, p_{m}$ where $\pi_i > \pi_j$ for $i < j$,  \ie $f^* = \arg \min \Rum$.
  Then, it follows that $f^*$ is also a minimizer of the true AUC risk $\Rpn$,  and thus, $\Rum$ is consistent with $\Rpn$.
\end{theorem}
Similarly, the desired model can be obtained by optimizing the U\(^m\) AUC risk minimization problem under the condition that only impure data sets are available. This means that the ERM in \cref{eq:msets} provides a naive solution for U\(^m\) AUC risk minimization by solving all the \(m(m-1)/2\) AUC sub-problems using a pairwise surrogate loss based on the definition of the AUC score \cite{consistency}, and minimizing the loss over all instance pairs that belong to different U sets. 

However, such a solution can be complex and inefficient, especially when dealing with a large number of datasets and a huge amount of data in each dataset.
For instance, with $m$ datasets, we need to handle $m(m-1)/2$ sub-problems according to the definition of U\(^m\) AUC risk, which can be complex when $m$ grows.
Furthermore, assuming that the number of samples is $n$, if a pairwise loss is used to optimize each sub-problem, the time complexity of each epoch in training is $O(n^2)$. This means that the time consumption of the method grows quadratically with $n$, making it computationally infeasible for large-scale datasets. To address the aforementioned issues, we propose a novel and efficient training algorithm for U\(^m\) AUC risk minimization.

\subsection{U{$^m$}-AUC: Simple and Efficient Learning for U{$^m$} AUC Risk Minimization}

To simplify the form of naive solution to U\(^m\) AUC risk minimization, we transform it into an equivalent multi-label learning problem, reducing the number of sub-problems to $m-1$. To decrease the time cost of training the model, we use an efficient stochastic optimization algorithm, reducing the time complexity from $O(n^2)$ to $O(n)$. The proposed approach is demonstrated in the \cref{fig:framework}.

\subsubsection{Reduction in the Number of Sub-problems}

To reduce the number of sub-problems, we transform the U\(^m\) AUC risk minimization problem into a multi-label learning problem with $m-1$ labels. We let the samples in datasets $\text{U}_1, \cdots, \text{U}_k$ have label $1$ at the $k$-th position and let the samples in datasets $\text{U}_{k+1}, \cdots, \text{U}_{m}$ have label $0$ at the $k$-th position.

Specifically, we assign the surrogate label \(\bar{\boldsymbol{y}}^{(k)}\) be the label of the \(k\)-th unlabeled set \(\text{U}_k\), where 
\begin{equation}
  \bar{\boldsymbol{y}}^{(k)} = [\, \underbrace{0,0,\ldots,0}_{k-1},\underbrace{1,1,\ldots,1}_{m-k}\,]
\end{equation}
has $k-1$ negative labels in
the front and \(m-k\) positive labels in the rear.

Let $\boldsymbol{g}(x) = \hat{\boldsymbol{y}}$ be the model output score for the multi-label learning problem, and $\boldsymbol{g}_k(x)$ be the $k$-th dimension of $\boldsymbol{g}(x)$, which denote the output of $k$-th sub-problem, 
the multi-label learning problem can be formalized as:
\begin{equation}\label{multi_label}
    \max_{\boldsymbol{g}} \quad \text{AUC}_\text{macro}(\boldsymbol{g}) = \frac{1}{m-1}\sum_{k=1,2,\cdots,m} \text{AUC}_k(\boldsymbol{g}_k) \,,
\end{equation}
and $\text{AUC}_k$ is AUC on the $k$-th label, which has
\begin{equation}
    \text{AUC}_k(\boldsymbol{g}_k) = 1 - \sum_{\xp \in \bigcup_{i\le k}{\text{U}_i}}\sum_{\xn \in \bigcup_{j > k}\text{U}_j}\frac{\ell(\boldsymbol{g}_k(\xp, \xn))}{\sum_{i\le k}n_i \sum_{j > k}n_j} \,.
\end{equation}

For the $k$-th label, the sub-problem of the multi-label learning problem is a simple AUC optimization problem:
\begin{equation}\label{multi_label_sub-problem}
    \min_{\boldsymbol{g}_k} \quad \frac{1}{m-1}\sum_{\xp \in \bigcup_{i\le k}{\text{U}_i}}\sum_{\xn \in \bigcup_{j > k}\text{U}_j}\frac{\ell(\boldsymbol{g}_k(\xp, \xn))}{\sum_{i\le k}n_i \sum_{j > k}n_j} \,.
\end{equation}
That is, in order to solve the multi-label learning problem described in~\cref{multi_label}, we can optimize $m-1$ sub-problems of the form described in~\cref{multi_label_sub-problem} only. The following explanation outlines why the \umauc{} problem~\cref{eq:rmsets} can be solved by solving this multi-label learning problem~\cref{multi_label}.

Let $r_{ijk} = (n_in_j)/(\sum_{i\le k}n_i \sum_{j > k}n_j)$, the optimization problem~\cref{multi_label_sub-problem} is equivalent to
\begin{equation}
    \min_{\boldsymbol{g}_k} \quad \frac{1}{m-1}\sum_{i\le k}\sum_{j>k} r_{ijk} \sum_{\xp \in \text{U}_i}\sum_{\xn \in \text{U}_j}\frac{\ell(\boldsymbol{g}_k(\xp, \xn))}{n_i n_j} \,,
\end{equation}
or simplified as:
\begin{equation}
    \min_{\boldsymbol{g}_k} \quad  \frac{1}{m-1}\sum_{i\le k}\sum_{j>k} r_{ijk} \empRij(\boldsymbol{g}_k) \,.
\end{equation}

This is exactly the special case of \cref{eq:msets}'s where \(z_{ij}=r_{ijk}/(m-1) > 0\). Therefore, each sub-problem of multi-label learning problem~\cref{multi_label} is an ERM problem for the U\(^m\) AUC risk minimization problem~\cref{eq:rmsets}. According to \cref{thm:con_msets}, optimizing this multi-label learning problem is equivalent to solving the original AUC optimization problem. Thus, we aggregate the output of sub-problem  as $f = \frac{1}{m-1}\sum_{k = 1}^{m-1}  \boldsymbol{g}_k$.

In summary, by transforming the U\(^m\) AUC risk minimization problem into a multi-label learning problem, we only need to optimize $m-1$, rather than ${m(m-1)}/{2}$ sub-problems as in the naive approach.

\subsubsection{Efficient Training of the Model}

Although we have reduced the number of the sub-problems from $O(m^2)$ to $O(m)$, this approach may not be practical for large datasets when optimizing a generic pairwise loss on training data, since the pairwise method suffers from severe scalability issue, as each epoch will take $O(n_P \cdot n_N)$ time with $n_P$ positive and $n_N$ negative samples. This issue has been discussed and efficient methods for AUC optimization have been proposed in several previous works like~\cite{Yuan_2021_ICCV,Liu2020}. These method will take only $O(n_P + n_N)$ time each epoch, making them more suitable for large-scale datasets.

Consider using the square surrogate AUC loss in the multi-label learning problem~\cref{multi_label}, with the derivation process shown in Appendix, we get
\begin{equation}\label{eq:obj}
    \begin{aligned}
    \ &\frac{1}{m-1}\sum_{1\le k < m}\sum_{\xp \in \bigcup_{i\le k}{\text{U}_i}}\sum_{\xn \in \bigcup_{j > k}\text{U}_j}\frac{(1-f(\xp)+f(\xn))^2}{\sum_{i\le k}n_i \sum_{j > k}n_j} \\
    =&\  \frac{1}{m-1}\sum_{1\le k < m}(\underbrace{\sum_{\xp \in \bigcup_{i\le k}{\text{U}_i}}\frac{(f(\xp)-a_k(f))^2}{\sum_{i\le k}n_i }}_{A_k(f)} \\
    &  + \underbrace{\sum_{\xn \in \bigcup_{j > k}\text{U}_j}\frac{(f(\xn)-b_k(f)))^2}{\sum_{j > k}n_j}}_{B_k(f)} 
     + \underbrace{(1-a_k(f)+b_k(f))^2}_{C_k(f)} )\\
    =&\ \frac{1}{m-1}\sum_{1\le k < m}(A_k(f) + B_k(f)\\
    &\quad \quad \quad \quad \quad + \max_{\alpha_k}\{2\alpha_k(1-a_k(f)+b_k(f))-\alpha_k^2\})
    \end{aligned}
\end{equation}
where $a_k(f) = \sum_{\xp \in \bigcup_{i\le k}{\text{U}_i}}f(\xp)/ {\sum_{i\le k}n_i}$, and $b_k(f) = \sum_{\xn \in \bigcup_{j > k}{\text{U}_j}}f(\xn)/ {\sum_{j > k}n_j}$.

Following previous work~\cite{solam}, the objective~\cref{eq:obj} is equivalent to $(m-1)$ min-max problems
\begin{equation}\label{min-max problem}
    \min_{f,a_k,b_k} \max_{\alpha_k} h(f,a_k,b_k,\alpha_k) := \mathbb{E}_{\boldsymbol{z}}[H(f,a_k,b_k,\alpha_k;\boldsymbol{z})] ,
\end{equation}
where $\boldsymbol{z}=(\xp,y)$ is a random sample, and
\begin{equation}
\begin{aligned}
    H(f,a_k,b_k,\alpha_k;\boldsymbol{z}) 
    = & (1-p)(f(\xp)-a_k)^2\mathbb{I}_{[y=1]} \\
    & + p(f(x)-b_k)^2\mathbb{I}_{[y=-1]}\\
    & - p(1-p)\alpha_k^2 \\
    & + 2\alpha_k(p(1-p)+pf(\xp)\mathbb{I}_{[y=-1]}\\
    &\quad \quad \quad \quad  -(1-p)f(\xp)\mathbb{I}_{[y=1]}),
\end{aligned}
\end{equation}
with $p =\sum_{i\le k}n_i / (\sum_{i\le k}n_i + \sum_{j > k}n_j)$.

Besides, we replace the $C_k$ with $\max_{\alpha_k \ge 0}\{2\alpha_k(m-a_k(f)+b_k(f))-\alpha_k^2\}$, which has a margin parameter $m$ to make the loss function more robust~\cite{Yuan_2021_ICCV}.

The min-max problems can be efficiently solved using primal-dual stochastic optimization techniques, eliminating the need for explicit construction of positive-negative pairs. In our implementation, we leverage the Polyak-\L ojasiewicz (PL) conditions~\cite{guo2020fast} of the objective functions in the min-max problems \cref{min-max problem}, and update the parameters accordingly to solve the multi-label learning problem.

Through a combination of equivalence problem conversion techniques and efficient optimization methods, the complexity of each epoch in training can be reduced from $O(n^2)$ to $O(n)$.
The algorithm is described in \cref{alg:umauc}.

\begin{algorithm}[t]
\caption{\umauc{}}
 \label{alg:umauc}
\hspace*{0.02in} {\bf Input:}
Model $\boldsymbol{g}$, $m$ sets of unlabeled data with class priors in descending order $U_1, \cdots, U_m$,  training epochs $\textit{num\_epochs}$, number of batches $\textit{num\_batchs}$.
\begin{algorithmic}[1]
\FOR{$t = 1, 2,\ldots, \textit{num\_epochs}$}
    \FOR{$b = 1, 2,\ldots, \textit{num\_batches}$}
        \STATE Fetch mini-batch $\mathcal{B}$ from $\bigcup_{0\le i\le m}{\text{U}_i}$
        \STATE Forward $\mathcal{B}$ and get $\boldsymbol{g}(\mathcal{B})$
        \STATE Compute multi-label loss of the mini-batch $\mathcal{B}$
        \STATE Update the parameters of $\boldsymbol{g}$ 
    \ENDFOR
\ENDFOR
\STATE Aggregate the $\boldsymbol{g}$ by $f = \frac{1}{m-1}\sum_{k = 1}^{m-1}  \boldsymbol{g}_k$
\end{algorithmic}
\hspace*{0.02in} {\bf Output: $f$}
\end{algorithm}

\subsection{Theoretical Analysis}

In this subsection, we provide a theoretical analysis of the approach described above. Specifically, we prove excess risk bounds for the ERM problem of U$^2$ AUC and U$^m$ AUC.

Let $\mathcal{X}$ be the feature space, $K$ be a kernel over $\mathcal{X}^2$, and $C_w$ be a strictly positive real number. Let $\mathcal{F}_{K}$ be a class of functions defined as:
$$
\mathcal{F}_{K} = \{f_w:\mathcal{X}\to R,f_w(x)=K(w,x)|\|w\|_k\leq  C_w \}\,,
$$
where $\|x\|_K=\sqrt{K(x,x)}$.
We also assume that the surrogate loss $\ell$ is $L$-Lipschitz continuous, bounded by a strictly positive real number $C_{\ell}$,
and satisfies inequality $\ell \geq \ell_{01}$.

Let $\hat f^*_{ij}$ be the minimizer of empirical risk $\empRij(f)$, 
we introduce the following excess risk bound, showing that the risk of $\hat f^*_{ij}$
converges to risk of the optimal function in the function family $\mathcal{F}_{K}$.

\begin{theorem}[Excess Risk for U\(^2\) AUC ERM problem]\label{thm:u2}
    Assume that $\hat f^*_{ij} \in \mathcal{F}_{K}$ is the minimizer of empirical risk $\empRij(f)$,
    $f^*_{PN} \in \mathcal{F}_{K}$ is the minimizer of true risk $\Rpn (f)$.
    For any $\delta>0$, with the probability at least $1-\delta$, we have
    $$
    \Rpn (\hat  f^*_{ij})-\Rpn (f^*_{PN}) \leq \frac{h(\delta)}{a}\sqrt{\frac{n_i+n_j}{n_i n_j}}\,,
    $$
    where $h(\delta)=8\sqrt{2}C_\ell C_w C_x+5\sqrt{2\ln{(2/\delta)}}$, $a = \pi_i - \pi_j $, and $n_i, n_j$ is the size of dataset $\text{U}_i, \text{U}_j$.

\end{theorem}

Theorem \ref{thm:u2} guarantees that the excess risk of general case can be bounded plus the term of order
$$
\mathcal{O}\left(\frac{1}{a \sqrt {n_{i}}}+\frac{1}{a \sqrt {n_{j}}}\right)\,.
$$

Let $\hat f^*_{U^m}$ be the minimizer of empirical risk $\empRum(f)$, 
we introduce the following excess risk bound, showing that the risk of $\hat f^*_{U^m}$
converges to risk of the optimal function in the function family $\mathcal{F}_{K}$.

\begin{theorem}[Excess Risk for U\(^m\) AUC ERM problem]\label{thm:um}
    Assume that $\hat f^*_{U^m} \in \mathcal{F}_{K}$ is the minimizer of empirical risk $\empRum(f)$,
    $f^*_{PN} \in \mathcal{F}_{K}$ is the minimizer of true risk $\Rpn (f)$.
    For any $\delta>0$, with the probability at least $1-\delta$, we have
    $$
    \begin{aligned}
    &\Rpn (\hat  f^*_{U^m})-\Rpn (f^*_{PN}) \leq \\
    &\quad \quad \quad \quad \frac{h(\frac{2\delta}{m(m-1)})}{s}\sum_{i,j|1\le i < j \le m}z_{ij}\sqrt{\frac{n_i+n_j}{n_i n_j}}\,,
    \end{aligned}
    $$
    where $h(\delta)=8\sqrt{2}C_\ell C_w C_x+5\sqrt{2\ln{(2/\delta)}}$, and $s = \sum_{i,j|1\le i < j \le m} z_{ij}(\pi_i-\pi_j)$ , $n_i, n_j$ is the size of unlabeled dataset $\text{U}_i, \text{U}_j$.

\end{theorem}

Theorem \ref{thm:um} guarantees that the excess risk of general case can be bounded plus the term of order
$$
\mathcal{O}\left(\frac{1}{s}\sum_{i,j|1\le i < j \le m} z_{ij}\sqrt{\frac{n_i+n_j}{n_i n_j}}\right)\,.
$$

It is evident that \cref{thm:um} degenerates into \cref{thm:u2} when $m = 2$ and $z_{12}=1$.

\section{Experiments}\label{sec:exp}

\begin{table*}[t]
\centering
\begin{tabular}{@{}c|c|cccc|c@{}}
\toprule
Dataset                    & $\mathcal{D}$      &    LLP-VAT$^\star$   &        LLP-VAT       &   U$^m$-SSC$^\star$  &       U$^m$-SSC      &          U$^m$-AUC           \\ 
\midrule
\multirow{4}{*}{K-MNIST\ ($m=10$)}    & $\mathcal{D}_{u}$  & $0.865_{\pm 0.0145}$ & $0.896_{\pm 0.0249}$ & $0.908_{\pm 0.0073}$ & $0.911_{\pm 0.0084}$ & $\mathbf{0.938}_{\pm 0.0064}$ \\
                           & $\mathcal{D}_{b}$  & $0.780_{\pm 0.0225}$ & $0.789_{\pm 0.0185}$ & $0.833_{\pm 0.0357}$ & $0.836_{\pm 0.0521}$ & $\mathbf{0.851}_{\pm 0.0616}$ \\
                           & $\mathcal{D}_{c}$  & $0.853_{\pm 0.0330}$ & $0.808_{\pm 0.0131}$ & $0.858_{\pm 0.0239}$ & $0.856_{\pm 0.0307}$ & $\mathbf{0.870}_{\pm 0.0512}$ \\
                           & $\mathcal{D}_{bc}$ & $0.825_{\pm 0.0315}$ & $0.798_{\pm 0.0332}$ & $0.868_{\pm 0.0255}$ & $0.857_{\pm 0.0390}$ & $\mathbf{0.896}_{\pm 0.0439}$ \\
\midrule
\multirow{4}{*}{CIFAR-10\ ($m=10$)}  & $\mathcal{D}_{u}$  & $0.856_{\pm 0.0131}$ & $0.856_{\pm 0.0066}$ & $0.860_{\pm 0.0090}$ & $0.859_{\pm 0.0131}$ & $\mathbf{0.905}_{\pm 0.0080}$ \\
                           & $\mathcal{D}_{b}$  & $0.723_{\pm 0.0454}$ & $0.737_{\pm 0.0754}$ & $0.746_{\pm 0.0614}$ & $0.778_{\pm 0.0462}$ & $\mathbf{0.866}_{\pm 0.0238}$ \\
                           & $\mathcal{D}_{c}$  & $0.787_{\pm 0.0172}$ & $0.847_{\pm 0.0059}$ & $0.792_{\pm 0.0372}$ & $0.807_{\pm 0.0209}$ & $\mathbf{0.884}_{\pm 0.0046}$ \\
                           & $\mathcal{D}_{bc}$ & $0.769_{\pm 0.0373}$ & $0.805_{\pm 0.0231}$ & $0.796_{\pm 0.0552}$ & $0.812_{\pm 0.0430}$ & $\mathbf{0.887}_{\pm 0.0155}$ \\
\midrule
\multirow{4}{*}{CIFAR-100\ ($m=10$)} & $\mathcal{D}_{u}$  & $0.734_{\pm 0.0092}$ & $0.731_{\pm 0.0167}$ & $0.747_{\pm 0.0192}$ & $0.756_{\pm 0.0115}$ & $\mathbf{0.847}_{\pm 0.0121}$ \\
                           & $\mathcal{D}_{b}$  & $0.630_{\pm 0.0183}$ & $0.651_{\pm 0.0210}$ & $0.652_{\pm 0.0332}$ & $0.667_{\pm 0.0331}$ & $\mathbf{0.715}_{\pm 0.0292}$ \\
                           & $\mathcal{D}_{c}$  & $0.670_{\pm 0.0168}$ & $0.707_{\pm 0.0117}$ & $0.676_{\pm 0.0363}$ & $0.692_{\pm 0.0264}$ & $\mathbf{0.757}_{\pm 0.0136}$ \\
                           & $\mathcal{D}_{bc}$ & $0.672_{\pm 0.0359}$ & $0.700_{\pm 0.0324}$ & $0.683_{\pm 0.0500}$ & $0.701_{\pm 0.0415}$ & $\mathbf{0.751}_{\pm 0.0641}$ \\

\midrule\midrule
\multirow{4}{*}{K-MNIST\ ($m=50$)}   & $\mathcal{D}_{u}$  & $0.896_{\pm 0.0124}$ & $0.902_{\pm 0.0102}$ & $0.915_{\pm 0.0136}$ & $0.915_{\pm 0.0107}$ & $\mathbf{0.931}_{\pm 0.0156}$ \\
                           & $\mathcal{D}_{b}$  & $0.808_{\pm 0.0142}$ & $0.787_{\pm 0.0196}$ & $0.861_{\pm 0.0102}$ & $0.869_{\pm 0.0083}$ & $\mathbf{0.883}_{\pm 0.0229}$ \\
                           & $\mathcal{D}_{c}$  & $0.863_{\pm 0.0206}$ & $0.833_{\pm 0.0165}$ & $0.855_{\pm 0.0378}$ & $0.863_{\pm 0.0417}$ & $\mathbf{0.867}_{\pm 0.0125}$ \\
                           & $\mathcal{D}_{bc}$ & $0.860_{\pm 0.0523}$ & $0.815_{\pm 0.0052}$ & $0.881_{\pm 0.0056}$ & $0.885_{\pm 0.0078}$ & $\mathbf{0.904}_{\pm 0.0012}$ \\
\midrule
\multirow{4}{*}{CIFAR-10\ ($m=50$)}  & $\mathcal{D}_{u}$  & $0.852_{\pm 0.0079}$ & $0.857_{\pm 0.0073}$ & $0.853_{\pm 0.0030}$ & $0.854_{\pm 0.0492}$ & $\mathbf{0.889}_{\pm 0.0083}$ \\
                           & $\mathcal{D}_{b}$  & $0.757_{\pm 0.0250}$ & $0.742_{\pm 0.0847}$ & $0.794_{\pm 0.0278}$ & $0.806_{\pm 0.0204}$ & $\mathbf{0.861}_{\pm 0.0097}$ \\
                           & $\mathcal{D}_{c}$  & $0.790_{\pm 0.0132}$ & $0.852_{\pm 0.0038}$ & $0.807_{\pm 0.0101}$ & $0.808_{\pm 0.0062}$ & $\mathbf{0.861}_{\pm 0.0138}$ \\
                           & $\mathcal{D}_{bc}$ & $0.804_{\pm 0.0056}$ & $0.830_{\pm 0.0235}$ & $0.826_{\pm 0.0059}$ & $0.832_{\pm 0.0052}$ & $\mathbf{0.873}_{\pm 0.0074}$ \\
\midrule
\multirow{4}{*}{CIFAR-100\ ($m=50$)} & $\mathcal{D}_{u}$  & $0.739_{\pm 0.0036}$ & $0.738_{\pm 0.0084}$ & $0.742_{\pm 0.0647}$ & $0.744_{\pm 0.0084}$ & $\mathbf{0.844}_{\pm 0.0042}$ \\
                           & $\mathcal{D}_{b}$  & $0.669_{\pm 0.0199}$ & $0.673_{\pm 0.0363}$ & $0.686_{\pm 0.0103}$ & $0.696_{\pm 0.0068}$ & $\mathbf{0.756}_{\pm 0.0281}$ \\
                           & $\mathcal{D}_{c}$  & $0.689_{\pm 0.0075}$ & $0.724_{\pm 0.0065}$ & $0.700_{\pm 0.0018}$ & $0.703_{\pm 0.0097}$ & $\mathbf{0.790}_{\pm 0.0085}$ \\
                           & $\mathcal{D}_{bc}$ & $0.699_{\pm 0.0065}$ & $0.718_{\pm 0.0082}$ & $0.714_{\pm 0.0009}$ & $0.717_{\pm 0.0024}$ & $\mathbf{0.812}_{\pm 0.0163}$ \\
\bottomrule
\end{tabular}
\caption{Test AUC on benchmark datasets and different priors distribution with $m=10$ and $m=50$.}
\label{tab:tabapp}
\end{table*}
 
In this section, we report the experimental results of the proposed \umauc{}, compared to state-of-the-art \um{} classification approaches.

\paragraph{Datasets}
We tested the performance of \umauc{} using the benchmark datasets Kuzushiji-MNIST (K-MNIST for short)~\cite{clanuwat2018deep}, CIFAR-10, and CIFAR-100~\cite{krizhevsky2009learning} with synthesizing multiple datasets with different settings. We transformed these datasets into binary classification datasets, where we classified odd vs. even class IDs for K-MNIST and animals vs. non-animals for CIFAR datasets.
In the experiments, we choose $m \in \{10,50\}$, and the size of each unlabeled data set $\text{U}_i$ is fixed to $n_i = \lceil n_{\text{train}}/m \rceil$, unless otherwise specified. To simulate the distribution of the dataset in different cases, we will generate the class priors $\{\pi_i\}_{i=1}^m$ from four different distributions , ensuring that the class priors are not all identical to avoid mathematically unsolvable situations. We then randomly sampled data from the training set into ${\text{U}_i}$ using the definition in \cref{umdef}. 
\paragraph{Models}
For all experiments on the Kuzushiji-MNIST dataset, we use a 5-layer MLP (multi-layer perceptron) as our model. For experiments on the CIFAR datasets, we use the Resnet32~\cite{he2016deep} as our model. 
We train all models for 150 epochs, and we report the AUC on the test set at the final epoch. 

\paragraph{Baselines}
In our experiments, we compared our method with state-of-the-art U$^m$ classification methods: LLP-VAT~\cite{tsai20} on behalf of EPRM methods, and U$^m$-SSC~\cite{Lu2021umssc} on behalf of ERM methods. Note that the previous methods require exact class priors, while in our setting, we can only obtain relative order relations for the class priors of the unlabeled dataset. To ensure fairness in performance comparisons, we created weaker versions of LLP-VAT and U$^m$-SSC, called LLP-VAT$^\star$ and U$^m$-SSC$^\star$, respectively, by giving them priors obtained by dividing $[0, 1]$ uniformly instead of using the true priors. We used Adam~\cite{kingma2014adam} and cross-entropy loss for their optimization, following the standard implementation in the original paper. To ensure fairness, we used the same model to implement all methods in all tasks. 
 
 Our implementation is based on PyTorch~\cite{NEURIPS2019_9015}, and experiments are conducted on an NVIDIA Tesla V100 GPU. To ensure the robustness of the results, all experiments are repeated 3 times with different random seed, and we report the mean values with standard deviations. For more details about the experiment, please refer to Appendix.
 
\subsection{Comparison with Baseline Methods}

To compare our approach with the baseline methods, we conducted experiments on the three image datasets and two different numbers of bags, as described above. In real-world scenarios, the class priors of datasets often do not follow a uniform distribution. To better simulate real-world situations, we considered four different class prior distributions for each image dataset and for each number of bags: $Beta(1,1)$, $Beta(5,1)$, $Beta(5,5)$, and $Beta(5,2)$. We refer to these four distributions as uniform, biased, concentrated, and biased concentrated, respectively. These four distributions represent four distinct cases as follows:
\begin{enumerate}
    \item $\mathcal{D}_{u}$ (Uniform): the class priors are sampled from uniform distribution on $[0,1]$; 
    \item $\mathcal{D}_{b}$ (Biased): the class priors are sampled from the distribution concentrated on one side, \ie most sets have more positive samples than negative samples; 
    \item $\mathcal{D}_{c}$ (Concentrated): the class priors are sampled from the distribution concentrated around 0.5, \ie most sets have close proportions of positive and negative samples; 
    \item $\mathcal{D}_{bc}$ (Biased Concentrated): the class priors are sampled from the distribution concentrated around 0.8, \ie most sets have close proportions of positive and negative samples, and positive samples more than negative samples. 
\end{enumerate}

\begin{table*}[t]
\centering
\begin{tabular}{@{}c|c|cccc|c@{}}
\toprule
Dataset                    & $m$   &        $\tau=0.8$       &        $\tau=0.6$       &        $\tau=0.4$       &        $\tau=0.2$       &       Random         \\ 
\midrule
\multirow{2}{*}{K-MNIST}   & $10$  & $0.936_{\pm 0.0042}$ & $0.934_{\pm 0.0095}$ & $0.926_{\pm 0.0038}$ & $0.928_{\pm 0.0046}$ & $0.928_{\pm 0.0196}$ \\
                           & $50$  & $0.938_{\pm 0.0106}$ & $0.932_{\pm 0.0047}$ & $0.937_{\pm 0.0097}$ & $0.928_{\pm 0.0042}$ & $0.941_{\pm 0.0186}$ \\
\midrule
\multirow{2}{*}{CIFAR-10}  & $10$  & $0.907_{\pm 0.0087}$ & $0.901_{\pm 0.0053}$ & $0.901_{\pm 0.0039}$ & $0.895_{\pm 0.0026}$ & $0.904_{\pm 0.0123}$ \\
                           & $50$  & $0.900_{\pm 0.0022}$ & $0.895_{\pm 0.0080}$ & $0.893_{\pm 0.0023}$ & $0.890_{\pm 0.0147}$ & $0.902_{\pm 0.0048}$ \\
\midrule
\multirow{2}{*}{CIFAR-100} & $10$  & $0.842_{\pm 0.0098}$ & $0.835_{\pm 0.0036}$ & $0.827_{\pm 0.0228}$ & $0.817_{\pm 0.0243}$ & $0.803_{\pm 0.0366}$ \\
                           & $50$  & $0.795_{\pm 0.0090}$ & $0.805_{\pm 0.0067}$ & $0.785_{\pm 0.0210}$ & $0.777_{\pm 0.0213}$ & $0.811_{\pm 0.0125}$ \\
\bottomrule
\end{tabular}

\caption{Test AUC on benchmark datasets with different imbalanced setting.}
\label{tab:tab2}
\end{table*}

It is worth mentioning that our experiments encompass a broader range of settings compared to previous work. Specifically, the $Beta(1,1)$ distribution corresponds to a uniform distribution on the interval $[0,1]$, which is similar to the setting explored in \cite{Lu2021umssc}.

For $m=10$ and $m=50$, the results obtained from different datasets and varied distributions of class priors are reported in \cref{tab:tabapp}. 
The results demonstrates that our proposed method, U$^m$-AUC, outperforms the baselines, even when a smaller amount of information is utilized.

\subsection{Robustness to Imbalanced Datasets}

One of the most prevalent challenges in classification tasks is handling imbalanced datasets, where the number of samples in each class is unequal. In the U$^m$ setting, we also encounter imbalanced datasets. If the number of samples in each dataset is unequal, it can result in biased models that prioritize the larger datasets, while underperforming on the minority class.

To assess the robustness of our method against imbalanced datasets, we conducted tests using various settings. Specifically, we generated imbalanced datasets in two ways following the approach proposed in~\cite{Lu2021umssc}:
\begin{enumerate}
    \item Size Reduction: With reduce ratio $\tau$, randomly select $\lceil m/2 \rceil$ datasets, and change their size to $\lceil \tau \cdot (n_{\text{train}}/m) \rceil$.
    \item Random: Randomly sample dataset size $n_i$ from range $[0, n_{\text{train}}]$, such that $\sum_{i=1}^m n_i = n_{\text{train}}$.
\end{enumerate}

The test AUC of \umauc{} on with different imbalance datasets is presented in \cref{tab:tab2}. It indicates that our method is reasonably robust to the imbalance settings, as it exhibits a slow decline in test performance and a slow increase in test performance variance as the reduction ratio decreases.

\section{Related Works}\label{sec:rela}
\paragraph{U{$^m$} Classification}
U$^m$ Classification involves learning a classifier from $m (m \ge 2)$ unlabeled datasets, where we have limited information about each dataset, typically the class priors of each dataset.
The U$^m$ Classification setting is a case of \emph{weak supervised learning}~\cite{wsl}, and can be traced back to a classical problem of \emph{learning with label proportions} (LLP)~\cite{quadrianto2008estimating}.
There are three main categories of previous approaches to solving the U$^m$ classification problem: clustering-based approaches, EPRM (empirical proportion risk minimization)-based approaches, and ERM (empirical risk minimization)-based approaches. For clustering-based approaches, \citeay{xu2004maximum} and \citeay{krause2010discriminative} assumed that each cluster corresponds to a single class and applied discriminative clustering methods to solve the problem. For EPRM-based approaches, \citeay{yu2014learning} aimed to minimize the distance between the average predicted probabilities and the class priors for each dataset U$_i$ (i.e., empirical proportion risk), while \citeay{tsai20} introduced consistency regularization to the problem. For ERM-based approaches, \citeay{scott20} extended the U$^2$ classification problem by ensembling classifiers trained on all pairs of unlabeled sets, while \citeay{Lu2021umssc} tackled the problem through a surrogate set classification task.
However, all of the previous works on U$^m$ classification have required knowledge of the class priors
and are unable to address situations where only the relative order of the unlabeled sets' class priors is known.

\paragraph{AUC Optimization}
As a widely-used performance measure alongside classification accuracy, AUC has received great attention from researchers, especially for problems with imbalanced data. While the goal is to train models with better AUC, studies~\cite{cortes2003auc} have shown that algorithms that maximize model accuracy do not necessarily maximize the AUC score. Accordingly, numerous studies have been dedicated to directly optimizing the AUC for decades~\cite{yang2022auc}. 
To enable efficient optimization of AUC, \citeay{opauc} proposed an AUC optimization algorithm using a covariance matrix, while \citeay{solam} optimized the AUC optimize problem as a stochastic saddle point problem with stochastic gradient-based methods. For AUC optimization with deep neural models, \citeay{Liu2020} introduced deep AUC maximization based on a non-convex min-max problem, and \citeay{Yuan_2021_ICCV} proposed an end-to-end AUC optimization method that is robust to noisy and easy data.
Recently, there are also studies of partial-AUC and multi-class AUC optimization~\cite{yang2021learning, yang2021all, zhu2022auc,yao2022large}.
In addition to the algorithms, significant work has been conducted on the theoretical aspects. For example, \citeay{consistency} investigated the consistency of commonly used surrogate losses for AUC, while \citeay{aucgeb} and \citeay{usunier2005data} studied the generalization bounds of AUC optimization models. The research on AUC optimization has led to the development of numerous real-world applications, such as software build outcome prediction~\cite{ijcai18}, medical image classification~\cite{Yuan_2021_ICCV}, and molecular property prediction~\cite{wang2022advanced}.
Most recently, there has been a growing body of research on weakly supervised AUC optimization. For example, \citeay{puauc} and \citeay{Xie2018} studied semi-supervised AUC optimization, \citeay{Charoenphakdee2019} studied the properties of AUC optimization under label noise, 
and \citeay{xie2023weakly} offered a universal solution for AUC optimization in various weakly supervised scenarios.
However, to the best of our knowledge, there has been no investigation into AUC optimization for U$^m$ classification to date.

\section{Conclusion}\label{sec:con}
In this work, we investigate the challenge of constructing AUC optimization models from multiple unlabeled datasets. To address this problem, we propose \umauc{}, a novel AUC optimization method with both simplicity and efficiency. 
\umauc{} is the first solution for AUC optimization under the \um{} learning scenario and provides a solution for \um{} learning without exact knowledge of the class priors. 
Furthermore, theoretical analysis demonstrates the validity of \umauc{}, while empirical evaluation demonstrates that \umauc{} exhibits superiority and robustness compared to the state-of-the-art alternatives.

\bibliography{umauc, uniauc}

\begin{thebibliography}{37}
\providecommand{\natexlab}[1]{#1}

\bibitem[{Agarwal et~al.(2005)Agarwal, Graepel, Herbrich, {Har-Peled}, and Roth}]{aucgeb}
Agarwal, S.; Graepel, T.; Herbrich, R.; {Har-Peled}, S.; and Roth, D. 2005.
\newblock Generalization Bounds for the Area Under the {ROC} Curve.
\newblock \emph{Journal of Machine Learning Research}, 6: 393--425.

\bibitem[{Bekker and Davis(2020)}]{pulsurvey}
Bekker, J.; and Davis, J. 2020.
\newblock Learning from Positive and Unlabeled Data: A Survey.
\newblock \emph{Machine Learning}, 109(4): 719--760.

\bibitem[{Charoenphakdee, Lee, and Sugiyama(2019)}]{Charoenphakdee2019}
Charoenphakdee, N.; Lee, J.; and Sugiyama, M. 2019.
\newblock On Symmetric Losses for Learning from Corrupted Labels.
\newblock In \emph{Proceedings of 36th International Conference on Machine Learning}, 961--970.

\bibitem[{Clanuwat et~al.(2018)Clanuwat, Bober-Irizar, Kitamoto, Lamb, Yamamoto, and Ha}]{clanuwat2018deep}
Clanuwat, T.; Bober-Irizar, M.; Kitamoto, A.; Lamb, A.; Yamamoto, K.; and Ha, D. 2018.
\newblock Deep Learning for Classical {Japanese} Literature.
\newblock \emph{arXiv preprint arXiv:1812.01718}.

\bibitem[{Cortes and Mohri(2003)}]{cortes2003auc}
Cortes, C.; and Mohri, M. 2003.
\newblock {AUC} optimization vs. error rate minimization.
\newblock In \emph{Advances in Neural Information Processing Systems 16}.

\bibitem[{Gao et~al.(2013)Gao, Jin, Zhu, and Zhou}]{opauc}
Gao, W.; Jin, R.; Zhu, S.; and Zhou, Z.-H. 2013.
\newblock One-Pass {AUC} Optimization.
\newblock In \emph{Proceedings of 30th International Conference on Machine Learning}, 906--914.

\bibitem[{Gao and Zhou(2015)}]{consistency}
Gao, W.; and Zhou, Z.-H. 2015.
\newblock On the Consistency of {AUC} Pairwise Optimization.
\newblock In \emph{Proceedings of 24th International Joint Conference on Artificial Intelligence}, 939--945.

\bibitem[{Guo et~al.(2020)Guo, Yan, Yuan, and Yang}]{guo2020fast}
Guo, Z.; Yan, Y.; Yuan, Z.; and Yang, T. 2020.
\newblock Fast objective \& duality gap convergence for nonconvex-strongly-concave min-max problems.
\newblock \emph{arXiv preprint arXiv:2006.06889}.

\bibitem[{Han et~al.(2021)Han, Yao, Liu, Niu, Tsang, Kwok, and Sugiyama}]{Han2021nllsurvey}
Han, B.; Yao, Q.; Liu, T.; Niu, G.; Tsang, I.~W.; Kwok, J.~T.; and Sugiyama, M. 2021.
\newblock A Survey of Label-noise Representation Learning: Past, Present and Future.
\newblock \emph{arXiv:2011.04406}.

\bibitem[{Hanley and McNeil(1982)}]{auc}
Hanley, J.~A.; and McNeil, B.~J. 1982.
\newblock The Meaning and Use of the Area Under a Receiver Operating Characteristic ({ROC}) Curve.
\newblock \emph{Radiology}, 143(1): 29--36.

\bibitem[{He et~al.(2016)He, Zhang, Ren, and Sun}]{he2016deep}
He, K.; Zhang, X.; Ren, S.; and Sun, J. 2016.
\newblock Deep residual learning for image recognition.
\newblock In \emph{Proceedings of the 2016 IEEE Conference on Computer Vision and Pattern Recognition}, 770--778.

\bibitem[{Kingma and Ba(2014)}]{kingma2014adam}
Kingma, D.~P.; and Ba, J. 2014.
\newblock Adam: A method for stochastic optimization.
\newblock \emph{arXiv preprint arXiv:1412.6980}.

\bibitem[{Krause, Perona, and Gomes(2010)}]{krause2010discriminative}
Krause, A.; Perona, P.; and Gomes, R. 2010.
\newblock Discriminative clustering by regularized information maximization.
\newblock In \emph{Advances in Neural Information Processing Systems 23}.

\bibitem[{Krizhevsky, Hinton et~al.(2009)}]{krizhevsky2009learning}
Krizhevsky, A.; Hinton, G.; et~al. 2009.
\newblock Learning multiple layers of features from tiny images.

\bibitem[{Liu et~al.(2020)Liu, Yuan, Ying, and Yang}]{Liu2020}
Liu, M.; Yuan, Z.; Ying, Y.; and Yang, T. 2020.
\newblock Stochastic {AUC} Maximization with Deep Neural Networks.
\newblock In \emph{Proceedings of the International Conference on Learning Representations}.

\bibitem[{Lu et~al.(2021)Lu, Lei, Niu, Sato, and Sugiyama}]{Lu2021umssc}
Lu, N.; Lei, S.; Niu, G.; Sato, I.; and Sugiyama, M. 2021.
\newblock Binary Classification from Multiple Unlabeled Datasets via Surrogate Set Classification.
\newblock In \emph{Proceedings of the 38th International Conference on Machine Learning}, volume 139, 7134--7144.

\bibitem[{Paszke et~al.(2019)Paszke, Gross, Massa, Lerer, Bradbury, Chanan, Killeen, Lin, Gimelshein, Antiga, Desmaison, Kopf, Yang, DeVito, Raison, Tejani, Chilamkurthy, Steiner, Fang, Bai, and Chintala}]{NEURIPS2019_9015}
Paszke, A.; Gross, S.; Massa, F.; Lerer, A.; Bradbury, J.; Chanan, G.; Killeen, T.; Lin, Z.; Gimelshein, N.; Antiga, L.; Desmaison, A.; Kopf, A.; Yang, E.; DeVito, Z.; Raison, M.; Tejani, A.; Chilamkurthy, S.; Steiner, B.; Fang, L.; Bai, J.; and Chintala, S. 2019.
\newblock {PyTorch}: An Imperative Style, High-Performance Deep Learning Library.
\newblock In \emph{Advances in Neural Information Processing Systems 32}, 8024--8035.

\bibitem[{Quadrianto et~al.(2008)Quadrianto, Smola, Caetano, and Le}]{quadrianto2008estimating}
Quadrianto, N.; Smola, A.~J.; Caetano, T.~S.; and Le, Q.~V. 2008.
\newblock Estimating labels from label proportions.
\newblock In \emph{Proceedings of the 25th International Conference on Machine Learning}, 776--783.

\bibitem[{Sakai, Niu, and Sugiyama(2018)}]{puauc}
Sakai, T.; Niu, G.; and Sugiyama, M. 2018.
\newblock Semi-Supervised {AUC} Optimization Based on Positive-Unlabeled Learning.
\newblock \emph{Machine Learning}, 107: 767--794.

\bibitem[{Scott and Zhang(2020)}]{scott20}
Scott, C.; and Zhang, J. 2020.
\newblock Learning from Label Proportions: A Mutual Contamination Framework.
\newblock In \emph{Advances in Neural Information Processing Systems 35}, 22256--22267.

\bibitem[{Tsai and Lin(2020)}]{tsai20}
Tsai, K.-H.; and Lin, H.-T. 2020.
\newblock Learning from Label Proportions with Consistency Regularization.
\newblock In \emph{Proceedings of The 12th Asian Conference on Machine Learning}, volume 129, 513--528.

\bibitem[{Usunier, Amini, and Gallinari(2005)}]{usunier2005data}
Usunier, N.; Amini, M.-R.; and Gallinari, P. 2005.
\newblock A Data-dependent Generalisation Error Bound for the {AUC}.
\newblock In \emph{ICML'05 Workshop ROC Analysis in Machine Learning}.

\bibitem[{Wang et~al.(2022)Wang, Liu, Luo, Xu, Xie, Wang, Cai, Qi, Yuan, Yang et~al.}]{wang2022advanced}
Wang, Z.; Liu, M.; Luo, Y.; Xu, Z.; Xie, Y.; Wang, L.; Cai, L.; Qi, Q.; Yuan, Z.; Yang, T.; et~al. 2022.
\newblock Advanced graph and sequence neural networks for molecular property prediction and drug discovery.
\newblock \emph{Bioinformatics}, 38(9): 2579--2586.

\bibitem[{Xie and Li(2018{\natexlab{a}})}]{ijcai18}
Xie, Z.; and Li, M. 2018{\natexlab{a}}.
\newblock Cutting the Software Building Efforts in Continuous Integration by Semi-Supervised Online {AUC} Optimization.
\newblock In \emph{Proceedings of 27th International Joint Conference on Artificial Intelligence}, 2875--2881.

\bibitem[{Xie and Li(2018{\natexlab{b}})}]{Xie2018}
Xie, Z.; and Li, M. 2018{\natexlab{b}}.
\newblock Semi-Supervised {AUC} Optimization Without Guessing Labels of Unlabeled Data.
\newblock In \emph{Proceedings of 32nd {AAAI} Conference on Artificial Intelligence}, 4310--4317.

\bibitem[{Xie et~al.(2023)Xie, Liu, He, Li, and Zhou}]{xie2023weakly}
Xie, Z.; Liu, Y.; He, H.-Y.; Li, M.; and Zhou, Z.-H. 2023.
\newblock Weakly Supervised {AUC} Optimization: A Unified Partial {AUC} Approach.
\newblock \emph{arXiv preprint arXiv:2305.14258}.

\bibitem[{Xu et~al.(2004)Xu, Neufeld, Larson, and Schuurmans}]{xu2004maximum}
Xu, L.; Neufeld, J.; Larson, B.; and Schuurmans, D. 2004.
\newblock Maximum margin clustering.
\newblock In \emph{Advances in Neural Information Processing Systems 17}.

\bibitem[{Yang and Ying(2022)}]{yang2022auc}
Yang, T.; and Ying, Y. 2022.
\newblock {AUC} maximization in the era of big data and {AI}: A survey.
\newblock \emph{{ACM} Computing Surveys}, 55(8): 1--37.

\bibitem[{Yang et~al.(2021{\natexlab{a}})Yang, Xu, Bao, Cao, and Huang}]{yang2021learning}
Yang, Z.; Xu, Q.; Bao, S.; Cao, X.; and Huang, Q. 2021{\natexlab{a}}.
\newblock Learning with multiclass AUC: Theory and algorithms.
\newblock \emph{IEEE Transactions on Pattern Analysis and Machine Intelligence}, 44(11): 7747--7763.

\bibitem[{Yang et~al.(2021{\natexlab{b}})Yang, Xu, Bao, He, Cao, and Huang}]{yang2021all}
Yang, Z.; Xu, Q.; Bao, S.; He, Y.; Cao, X.; and Huang, Q. 2021{\natexlab{b}}.
\newblock When all we need is a piece of the pie: A generic framework for optimizing two-way partial AUC.
\newblock In \emph{International Conference on Machine Learning}, 11820--11829. PMLR.

\bibitem[{Yao, Lin, and Yang(2022)}]{yao2022large}
Yao, Y.; Lin, Q.; and Yang, T. 2022.
\newblock Large-scale optimization of partial auc in a range of false positive rates.
\newblock \emph{Advances in Neural Information Processing Systems}, 35: 31239--31253.

\bibitem[{Ying, Wen, and Lyu(2016)}]{solam}
Ying, Y.; Wen, L.; and Lyu, S. 2016.
\newblock Stochastic Online {AUC} Maximization.
\newblock In \emph{Advances in Neural Information Processing Systems 29}, 451--459.

\bibitem[{Yu et~al.(2014)Yu, Choromanski, Kumar, Jebara, and Chang}]{yu2014learning}
Yu, F.~X.; Choromanski, K.; Kumar, S.; Jebara, T.; and Chang, S.-F. 2014.
\newblock On learning from label proportions.
\newblock \emph{arXiv preprint arXiv:1402.5902}.

\bibitem[{Yuan et~al.(2021)Yuan, Yan, Sonka, and Yang}]{Yuan_2021_ICCV}
Yuan, Z.; Yan, Y.; Sonka, M.; and Yang, T. 2021.
\newblock Large-Scale Robust Deep {AUC} Maximization: A New Surrogate Loss and Empirical Studies on Medical Image Classification.
\newblock In \emph{Proceedings of the IEEE/CVF International Conference on Computer Vision (ICCV)}, 3040--3049.

\bibitem[{Zhou(2017)}]{wsl}
Zhou, Z.-H. 2017.
\newblock A brief introduction to weakly supervised learning.
\newblock \emph{National Science Review}, 5(1): 44--53.

\bibitem[{Zhu et~al.(2022)Zhu, Li, Wang, Wu, and Yang}]{zhu2022auc}
Zhu, D.; Li, G.; Wang, B.; Wu, X.; and Yang, T. 2022.
\newblock When AUC meets DRO: Optimizing partial AUC for deep learning with non-convex convergence guarantee.
\newblock In \emph{International Conference on Machine Learning}, 27548--27573. PMLR.

\bibitem[{Zhu et~al.(2009)Zhu, Goldberg, Brachman, and Dietterich}]{zhubook}
Zhu, X.; Goldberg, A.~B.; Brachman, R.; and Dietterich, T. 2009.
\newblock \emph{Introduction to Semi-Supervised Learning}.
\newblock Morgan and Claypool publishers.

\end{thebibliography}

\end{document}